\documentclass[final]{cvpr}

\usepackage{times}
\usepackage{epsfig}
\usepackage{graphicx}
\usepackage{amsmath}
\usepackage{amssymb}

\usepackage{multirow}
\usepackage{enumitem}
\def\gou{{$\checkmark$}}
\newcommand{\tb}[1]{\textbf{#1}}

\newcommand\blfootnote[1]{%
\begingroup 
\renewcommand\thefootnote{}\footnote{#1}%
\addtocounter{footnote}{-1}%
\endgroup 
}
\usepackage[pagebackref=true,breaklinks=true,colorlinks,bookmarks=false]{hyperref}

\begin{document}

\title{Improving Sign Language Translation with Monolingual Data\\by Sign Back-Translation}

\author{Hao~Zhou$^{1}$~~~~~Wengang~Zhou$^{1,2,}$\thanks{Corresponding authors: Wengang Zhou and Houqiang Li}~~~~~Weizhen~Qi$^{1}$~~~~~Junfu~Pu$^{1}$~~~~~Houqiang~Li$^{1,2,}$\footnotemark[1]\\
{\normalsize$^{1}$ CAS Key Laboratory of GIPAS, EEIS Department, University of Science and Technology of China}\\
{\normalsize$^{2}$ Institute of Artificial Intelligence, Hefei Comprehensive National Science Center}\\
{\tt\small zhouh156@mail.ustc.edu.cn, zhwg@ustc.edu.cn, \{weizhen,pjh\}@mail.ustc.edu.cn, lihq@ustc.edu.cn}
}

\maketitle
\thispagestyle{empty}
\pagestyle{empty}

\begin{abstract}
Despite existing pioneering works on sign language translation (SLT), there is a non-trivial obstacle, i.e., the limited quantity of parallel sign-text data. 
To tackle this parallel data bottleneck, we propose a sign back-translation (SignBT) approach, which incorporates massive spoken language texts into SLT training. 
With a text-to-gloss translation model, we first back-translate the monolingual text to its gloss sequence. 
Then, the paired sign sequence is generated by splicing pieces from an estimated gloss-to-sign bank at the feature level. 
Finally, the synthetic parallel data serves as a strong supplement for the end-to-end training of the encoder-decoder SLT framework. 

To promote the SLT research, we further contribute CSL-Daily, a large-scale continuous SLT dataset\blfootnote{\scriptsize{\url{http://home.ustc.edu.cn/~zhouh156/dataset/csl-daily}}}. 
It provides both spoken language translations and gloss-level annotations. 
The topic revolves around people's daily lives (e.g., travel, shopping, medical care), the most likely SLT application scenario.
Extensive experimental results and analysis of SLT methods are reported on CSL-Daily. 
With the proposed sign back-translation method, we obtain a substantial improvement over previous state-of-the-art SLT methods. 

\vspace{-4pt}
\end{abstract}

\section{Introduction}

Sign language serves as the primary communication method among the deaf community. 
However, in a society where the spoken language is primarily used, the deaf people face issues of social isolation and communication barrier in daily lives~\cite{bragg2019slp}. 
Due to the significant social impact and the cross-modality challenge, sign language understanding has been attracting more and more research attention~\cite{bsl1k-20,bragg2019slp,slt-nslt-cihan18,huang18han,hmm-koller-tpami19,islr-transfer-cvpr2020,fingerspelling19iccv,wang16isolated}. 
In this paper, we concentrate on sign language translation (SLT), 
which aims to automatically generate the spoken language translation from a continuous sign video. 

Considering the different grammar rules and vocabularies of sign language and spoken language, SLT is typically treated as a sequence-to-sequence learning problem. 
Existing SLT systems typically rely on the encoder-decoder architectures~\cite{slt-nslt-cihan18,slt-trans-cihan20,tspnet-nips20}. 
Despite the success of encoder-decoder networks in neural machine translation (NMT), the translation quality in SLT is limited, which is partially attributed to the huge gap in the training data size. 
While the News Translation task~\cite{WMT19findings} provides over 77M English-German data, 
the only suitable SLT dataset PHOENIX-2014T~\cite{slt-nslt-cihan18} has less than 9K Sign-German data. 
To alleviate it, there are two possible solutions, \ie, collecting millions of parallel pairs or introducing monolingual data. 
The high charge of sign video collection and annotation makes the former a luxury. 
In contrast, making good use of accessible monolingual texts is a promising direction for SLT.

\begin{figure}[tp]
   \centering
   \includegraphics[width=0.47\textwidth]{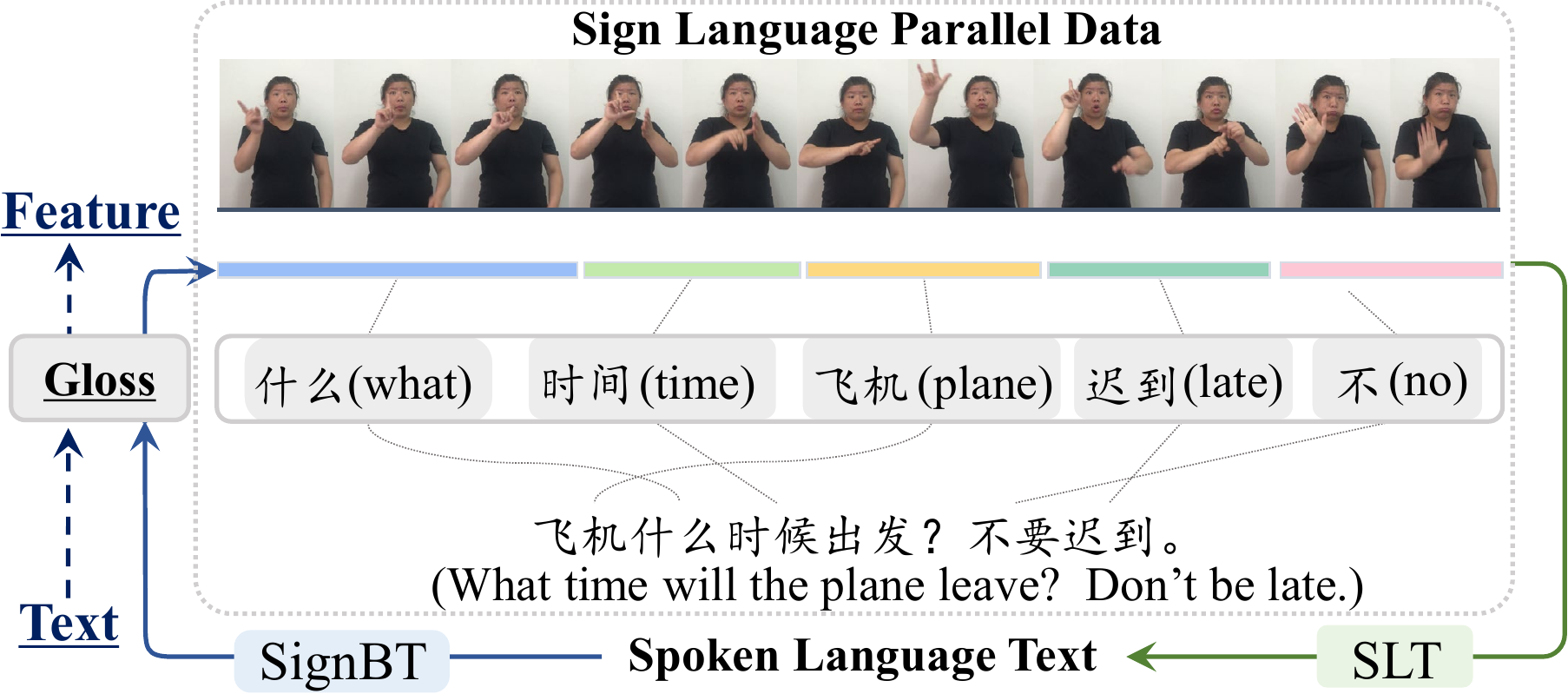}
   \caption{Pipelines of sign language translation (SLT) and sign back-translation (SignBT). 
   Our SignBT approach establishes an inverse path of SLT and uses it to enrich text-feature pairs from external monolingual data for SLT training.}\label{fig:pipeline}
   \vspace{-4pt}
\end{figure}

In this work, we propose to generate synthetic parallel data with monolingual texts for SLT training. 
Our method is inspired by the success of text-to-text back-translation in NMT~\cite{backtranslationACL16}. 
They train an inverse model with available pairs and use it to back-translate the monolingual data. 
However, when it goes to SLT, 
the key challenge becomes how to bridge the huge domain gap between text and vision signals. 
A straightforward idea is to generate the sign video from a sentence, which, 
however is a more challenging task involving various immature techniques, such as skeleton prediction~\cite{SLG-ECCV20}, gesture generation~\cite{SLG-TMM-THU-20} and temporal coherence fidelity~\cite{MonkeyNet-CVPR19}. 
A compromise option is to regress the feature sequence of video frames from a sentence. 
Unfortunately, it is an indeterminate problem and hard to formulate, because one sentence may correspond to numerous possible feature sequences, since the feature space of sign videos is far larger than the combination space of text vocabulary. 

To avoid the above problems, we propose a two-stage sign back-translation (SignBT) approach: text-to-gloss and gloss-to-sign. 
It is formulated as an inverse problem of SLT with an additional signal ``gloss'' (see Figure~\ref{fig:pipeline}). 
Gloss is a token of sign language word, 
which is annotated along with the order of signs in a video with no clear boundaries. 
We first train a text-to-gloss translator with available text-gloss pairs and predict the gloss sequence for each monolingual text. 
Then, to achieve the sequence-level gloss-to-sign conversion, we adopt a primitive but effective method, splicing sign pieces from features of segmented videos, which is somewhat analogous to concatenative text-to-speech synthesis~\cite{hunt1996unit,taylor2009text}. 
To acquire the precise boundary of each gloss, we train a sign-to-gloss network with connectionist temporal classification (CTC)~\cite{CTCLoss} and find the most likely alignment path for segmentation. 
The sign pieces could be segmented and stored as a sign bank in advance. 
Finally, we simplify the whole process into a text-to-text back-translation problem and a sequence splicing operation from pieces in the bank. 

The key reason that the synthetic data benefits SLT training lies in the two aspects of realism, 
\ie, the target text from the real language corpus and the source sign sequence spliced from the real feature bank. 
Though the fake pair may not be perfect as a real training data, it helps regularize the decoder when speaking target language and improve the robustness of extracting information from the source. 
Through extensive experiments, we verify the significant improvement of SLT models brought by monolingual data. 

Acquiring high-quality corpus is always crucial for SLT. 
In this paper, we provide the first large-scale Chinese Sign Language Translation benchmark, CSL-Daily. 
The native expression, compact annotation and clear hand details make our corpus suitable for a series of sign language research, 
\eg, sign language recognition, translation and generation. 
The evaluations on CSL-Daily of various SLT baselines are reported with in-depth analysis. 

Our main contributions are summarized as follows,
\begin{itemize}[itemsep=1pt,topsep=1pt,parsep=1pt,partopsep=0pt]
   \item We propose a sign back-translation approach to tackle parallel data shortage in SLT. 
   \item We contribute a new large-scale SLT benchmark with rich contents and compact annotations. 
   \item Extensive experiments on two datasets demonstrate the effectiveness of our SignBT mechanism. 
\end{itemize}

\begin{table*}[ht] \label{tab:dataset}
   \centering
   \footnotesize
   \caption{ Summary of public available video-based sign language benchmarks popular for computer vision research. 
         (SignDict: the corpus has isolated or segmented sign videos as a dictionary. Continuous: the corpus is composed of videos of continuous sign sentences and gloss-level annotations. Translation: the corpus has spoken language translation annotations.)} \label{tab:signdataset}
   \setlength{\tabcolsep}{4.2pt}{
   \begin{tabular}{l|c|cccc|crc|c}
   \hline
   \multirow{2}{*}{Dataset} & \multirow{2}{*}{Language} &\multicolumn{4}{c|}{Attribute}   &\multicolumn{3}{c|}{Statitics}& \multirow{2}{*}{Source} \\
          & &SignDict &Continuous &Translation  &Resolution & \#Signs    & \#Videos (avg. signs)  & \#Signers&          \\ \hline 
   DEVISIGN~\cite{wang16isolated}            &CSL&\gou &    &                &     -            &2,000  & 24,000 (1) & 8        &Lab   \\
   ICSL~\cite{csl-icme16}                    &CSL&\gou &    &                & 1280$\times$720  & 500   &125,000 (1) & 50       &Lab   \\ 
   MSASL~\cite{msasl-bmvc19}                 &ASL&\gou &    &                &     -            &1,000  & 25,513 (1) & 222      &Web   \\ 
   WLASL~\cite{wlasl-asu20}                  &ASL&\gou &    &                &     -            &2,000  & 21,083 (1) & 119      &Web   \\ 
   BSL-1K~\cite{bsl1k-20}                    &BSL&\gou &    &                &     -            &1,064  &273,000 (1) & 40       &TV    \\ 
   INCLUDE~\cite{include-20-ISL}             &ISL&\gou &    &                & 1920$\times$1080 &263    &  4,287 (1) &  7       &Lab    \\ \hline
   PHOENIX-2014~\cite{phoenix2014}           &DGS&     &\gou&                & 210$\times$260   &1,081  &6,841 (11) & 9         &TV    \\ 
   CCSL~\cite{huang18han}                    &CSL&\gou &\gou&                & 1280$\times$720 & 178   &25,000 (4) & 50        &Lab   \\ 
   SIGNUM~\cite{signum07}                    &DGS&\gou &\gou&\gou\ (German)  & 776$\times$578   & 455   &15,075 (7) & 25        &Lab  \\ \hline  
   PHOENIX-2014T~\cite{slt-nslt-cihan18}     &DGS&     &\gou&\gou\ (German)  & 210$\times$260   &1,066  &8,257 (9)  & 9         &TV    \\  
   \textbf{CSL-Daily} (ours)                 &CSL&\gou &\gou&\gou\ (Chinese) & 1920$\times$1080 &2,000  &20,654 (7) & 10        &Lab   \\ 
   \hline
   \end{tabular}
   }
\vspace{-8pt}
\end{table*}

\section{Related Work} \label{sec:related_work}
\textbf{Sign Language Recognition.} 
Sign language recognition (SLR) includes two sub-tasks: isolated SLR and continuous SLR. 
While isolated SLR aims to recognize one sign from a trimmed video, continuous SLR tries to recognize an ordered sign gloss sequence from a continuous video. 
Early works in isolated SLR utilize hand-crafted features~\cite{ong2005automatic,pami1998-sign} for sign description. 
With the success of deep learning, 2D and 3D convolutional neural networks (CNN)~\cite{I3D_Kinetics,yukai3d,vgg-two-stream} achieve favorable performance on action related tasks~\cite{I3D_Kinetics}. 
It inspires more research groups to study continuous SLR with large-scale vocabulary~\cite{staged,hmm-koller-tpami19,ian}.
To enable end-to-end training, connectionist temporal classification (CTC)~\cite{CTCLoss} is widely adopted by continuous SLR models~\cite{cslr-fcn-20fully,cui-tmm19,cslr-20-Stochastic,pu2020boosting,cslr-CTF-MM}. 
With the development of neural machine translation, Camgöz~\etal formulate a new task, neural sign language translation (SLT)~\cite{slt-nslt-cihan18}, which is becoming an active and promising direction~\cite{slt-trans-cihan20,tspnet-nips20}.

\textbf{Sign Language Translation.} 
SLT is different from SLR mainly in the aspect of sequence learning. 
The encoder-decoder based methods~\cite{seq2seq-attn-Bahdanau,luong-seq2seq,seq2seq-Sutskever} are adopted to process the difference in word order and vocabulary between sign language and spoken language. 
In~\cite{slt-nslt-cihan18}, Camgöz \etal propose an SLT dataset PHOENIX-2014T and provide spoken language annotations. 
They use attention-based encoder-decoder models to learn how to translate from spatial representations or sign glosses. 
Recently, transformer networks~\cite{vaswani2017attention} have been popular for neural machine translation (NMT). 
Camgöz \textit{et al.}~\cite{slt-trans-cihan20} apply transformers into sequence learning of sign language. Their work explores the multi-task formulation of continuous SLR and SLT. 
Under transformer framework, Li~\etal~\cite{tspnet-nips20} explore the hierarchical structure in sign video representation. 
Besides, some works~\cite{slt-camgoz2020multi,stmc-tmm} improve the SLT framework by considering the multi-cue characteristic of sign language. 

\textbf{Monolingual Data Exploration.} 
The integration of monolingual data for neural machine translation (NMT) models is first investigated in~\cite{gulcehre2015using}. 
Gulcehre ~\etal train the NMT model independently and use a language model from monolingual data for re-scoring during the decoding process. 
To introduce monolingual data in model training, Sennrich \etal propose a back-translation approach~\cite{backtranslationACL16} to generate synthetic parallel data for training without changing the encoder-decoder structure. 
In~\cite{chung2017lip}, sentences with blank facetracks are utilized to enhance the decoder of lip reading. 
Different from previous work, we design a sign back-translation mechanism across domains of video and language, which brings state-of-the-art results in SLT datasets and new insight to approach SLT. 

\textbf{Sign Language Dataset.} 
High-quality datasets are essential in promoting sign language research. 
A summary of publicly available datasets for video-based sign language research is presented in Table~\ref{tab:dataset}. 
The majority of them are composed of word-level sign videos. 
To achieve continuous SLR evaluation, some datasets provide gloss-level annotations~\cite{huang18han,phoenix2014,signum07}. 
Although German translations are provided in SIGNUM, it is not appropriate for SLT tasks, due to its limited vocabulary and sentences. 
Hence, PHOENIX-2014T~\cite{slt-nslt-cihan18} becomes the only suitable dataset for SLT research~\cite{slt-trans-cihan20,tspnet-nips20}.  
However, the frames in~\cite{phoenix2014} are cropped from a specific TV program of the weather forecast and thus in a low resolution. 
It constrains the exploration of language model and sign language generation in hand details. 
As a considerable complement, 
our CSL-Daily contains over 20K 1080P sign videos. 
It provides both gloss and translation annotations and covers diverse topics of daily lives.

\section{Proposed Method}
\begin{figure}[tp]
   \centering
   \includegraphics[width=0.47\textwidth]{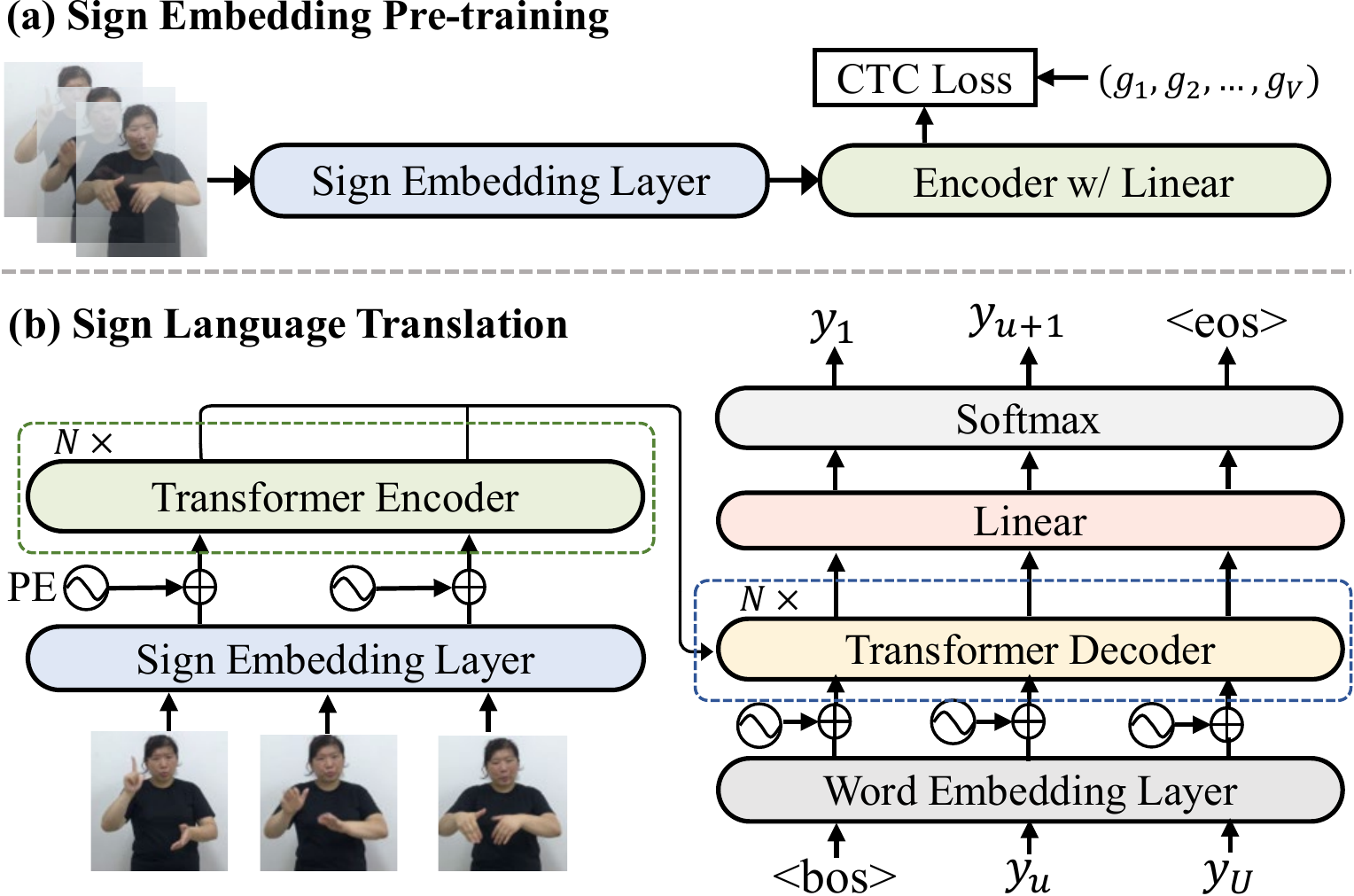}
   \caption{An overview of our SLT framework. We show the sign embedding pre-training process in (a). 
   It is trained with CTC Loss and gloss-level annotations. 
   The detailed encoder-decoder structure for SLT is shown in (b). (PE: Positional Encoding.)
   }\label{fig:framework}
\vspace{-8pt}
\end{figure}

Given a sign video $\mathbf{x}=\{x_t\}^T_{x=1}$ with $T$ Frames, 
sign language translation (SLT) can be formulated as learning the conditional probability $p(\mathbf{y}|\mathbf{x})$ of generating a spoken language sentence $\mathbf{y}=\{y_u\}^U_{u=1}$ with $U$ words. 
In addition, existing SLT datasets provide gloss-level annotations for pre-training of sign embedding networks.
Unlike spoken language which is non-monotonic to sign language,  
the gloss-level annotation $\mathbf{g}=\{g_v\}^V_{v=1}$ with $V$ sign glosses is order-consistent with sign gestures. 
An overview of our framework for SLT is depicted in Figure~\ref{fig:framework}. 

The remainder of this section is organized as follows. 
In subsection~\ref{subsec:signbank}, we elaborate the building method of our sign bank with a pre-trained sign embedding network. 
Then, the transformer-based SLT framework is revisited. Finally, we detail our sign back-translation process in subsection~\ref{subsec:signbt}. 

\subsection{Sign Bank Generation} \label{subsec:signbank}

To acquire the gloss-to-sign mapping, we are dedicated to build a sign bank containing video piece features indexed by its gloss vocabulary. 
However, due to the high cost of hiring experts, existing continuous sign datasets only have sentence-level gloss annotations~\cite{slt-nslt-cihan18,phoenix2014,csl-icme16} without the boundary ground-truth. 
It hinders the segmentation of sign feature sequences for our sign bank. 
Hence, we propose to establish the sign bank with estimated alignment paths from a pre-trained sign embedding network. 

\textbf{Sign Embedding Layer.} 
Unlike word embedding technique in NMT, which serves for word association learning, 
sign embedding is to convert a series of video frames to its feature representation. 
Our sign embedding layer $\Omega$ adopts a combination of 2D and 1D CNNs for spatiotemporal encoding~\cite{cui-tmm19}. 
In this work, the embedding operation is performed in the clip-level. 
We split video frames $\mathbf{x}$ into $N$ clips $\mathbf{c}=\{c_n\}_{n=1}^N$. 
The number of clips is $N=\lceil\frac{T}{s}\rceil$ with sliding window size $w$ and stride size $s$. 
By passing clips through $\Omega$, the embeddings $\mathbf{f}=\{f_n\}^N_{n=1}$ are extracted as follows, 
\begin{align} \label{eq:signembedding}
   f_n = \text{SignEmbedding}(c_n)=\Omega_\theta(c_n),
\end{align}
where $\theta$ denotes the parameters of the CNN network. 

\textbf{Sign-to-Gloss Pre-Training.}
The embedding layer is usually pre-trained with gloss-level annotations~\cite{slt-nslt-cihan18,slt-trans-cihan20}. 
For our embedding layer, we use connectionist temporal classification~\cite{CTCLoss} (CTC) with a transformer encoder~\cite{trans-vaswani17attention} for gloss-level temporal modelling.
The gloss probabilities $p(g_n|\mathbf{f})$ at the $n$-th time step could be estimated by a linear layer with softmax activation. 
According to CTC, the conditional probability $p(\mathbf{g} | \mathbf{x})$ is computed as the sum of probabilities of all feasible paths, 
\begin{align} \label{eq:gloss-predict}
   p(\mathbf{g} | \mathbf{x}) = \sum_{\pi \in \mathcal{B}^{-1}(\mathbf{g})}  p(\pi|\mathbf{f}),
\end{align}
where $\pi$ is a sign-to-gloss alignment path and $\mathcal{B}$ denotes the mapping between them. 
The embedding layer is trained through the CTC Loss $L_\text{ctc}=-\ln p(\mathbf{g}|\mathbf{x})$. 

\begin{figure}[tp]
   \centering
   \includegraphics[width=0.47\textwidth]{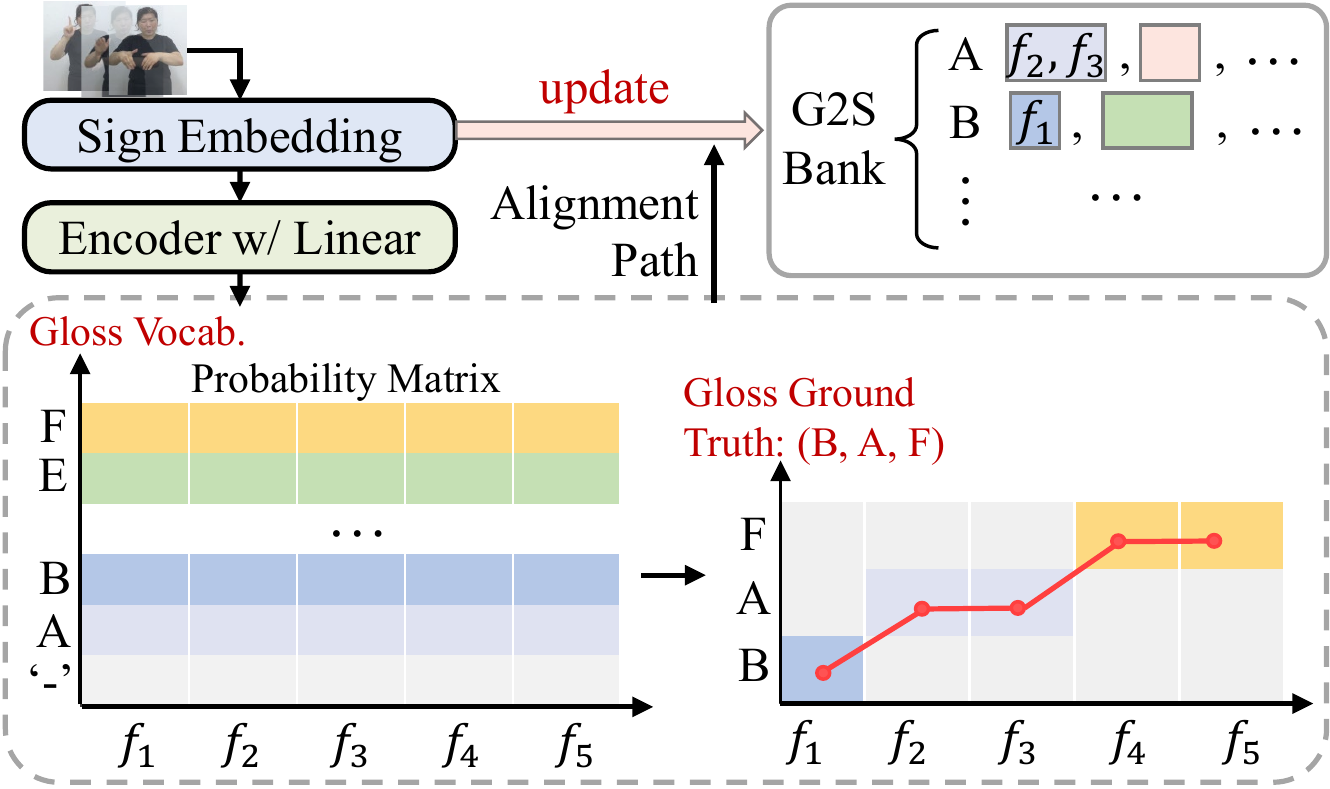}
   \caption{Illustration of constructing a Gloss-to-Sign (G2S) Bank according to the most probable alignment paths. 
   }\label{fig:G2Sbank}
\vspace{-8pt}
\end{figure}

\textbf{Gloss-to-Sign Bank.} 
Given a sign embedding sequence $\mathbf{f}=\{f_n\}_{n=1}^N$ extracted from $\Omega$ and its corresponding gloss sequence $\mathbf{g}=\{g_v\}_{v=1}^V$, 
we find the most probable alignment path $\hat{\pi}$ between them as follows, 
\begin{align} \label{eq:findpath}
   \hat{\pi} = \mathop{\arg\max}_{\pi \in \hat{\mathcal{B}}^{-1}(\mathbf{g})} p(\pi|\mathbf{f}).
\end{align}
The searching space is constrained within paths that conform to the mapping function $\hat{\mathcal{B}}$ with no blank labels (See Figure~\ref{fig:G2Sbank}). 
Notably, paths going through the blank label~\cite{CTCLoss} are excluded from the searching space to ensure the proper length as a sign sentence after splicing. 
The searching problem could be accelerated with Viterbi algorithm\cite{viterbi1967error,DPD_icme19}. 

With the estimated alignments, we segment the embedding sequence of each video into gloss pieces. 
They constitute a gloss-to-sign (G2S) bank in embedding space with a look-up table which is indexed by the gloss vocabulary. 
Each gloss slot may have multiple feature pieces. 

\subsection{SLT Training with Monolingual Data} \label{subsec:signbt}

Comparing with the limited size of sign-text pairs, the monolingual spoken language corpus is easy to reach the volume of millions. 
To make use of the monolingual texts, we propose to establish an inverse path of SLT and use it to enrich parallel data for training.

\textbf{Sign Language Translation.}
The encoder-decoder framework is widely utilized and explored in SLT~\cite{slt-nslt-cihan18,slt-trans-cihan20}. 
Here, we briefly introduce the transformer-based encoder-decoder structure in our SLT framework (see Figure~\ref{fig:framework}b). 
\textit{Notably}, our approach is not limited to this architecture. 

The encoder is composed of several stacked identical layers. 
Each layer has a self-attention network and a feed-forward network. 
To provide sequential clues, the input of the first layer is summed with a positional encoding (PE) vector as $\hat{f}_n = f_n + \text{PE}(n)$. 
The encoder takes all encoded input $\hat{\mathbf{f}}$ and generates N hidden vectors as follows, 
\begin{align}
   h_{1:N} &= \text{Encoder}(\hat{f}_{1:N}).
\end{align}

During decoding, we first pass each word $y_u$ through a lookup table for word embedding as follows, 
\begin{align}
   &w_u = \text{WordEmbedding}(y_u).  
\end{align}
Here, $\hat{w}_{u} = w_{u} + \text{PE}(u)$ is the positionally encoded word embedding of $y_u$. 
The decoder network includes an extra layer which performs attention operation over the encoder hidden vectors $h_{1:N}$ and hidden states of previously predicted words, for information aggregating. 
Then, the probability of words at the $u$-th step is generated as follows, 
\begin{align}
   &o_u = \text{Decoder}(\hat{w}_{1:u-1}, h_{1:N}), \label{eq:wordbyword}  \\ 
   &z_u = \text{softmax}(W o_u +b).
\end{align}
The initial word is $<\!bos\!>$ which indicates the beginning of a sentence. 
Finally, we compute the conditional probability of $p(\mathbf{y}|\mathbf{x})$ as follows, 
\begin{align} \label{eq:decode}
   p(\mathbf{y}|\mathbf{x}) = \prod_{u=1}^{U} p(y_u|y_{1:u-1}, \mathbf{x}) = \prod_{u=1}^{U} z_{u, y_u}.
\end{align}

To optimize the whole structure, the objective function is formulated as $L_{\text{SLT}} = -\ln p(\mathbf{y}|\mathbf{x})$. 
During inference, the words in spoken language text are predicted word-by-word as in Equation~\ref{eq:wordbyword}. 
The beam search~\cite{googletranslation} strategy is used to estimate a better decoding path in an acceptable range.

\begin{figure}[tp]
   \centering
   \includegraphics[width=0.47\textwidth]{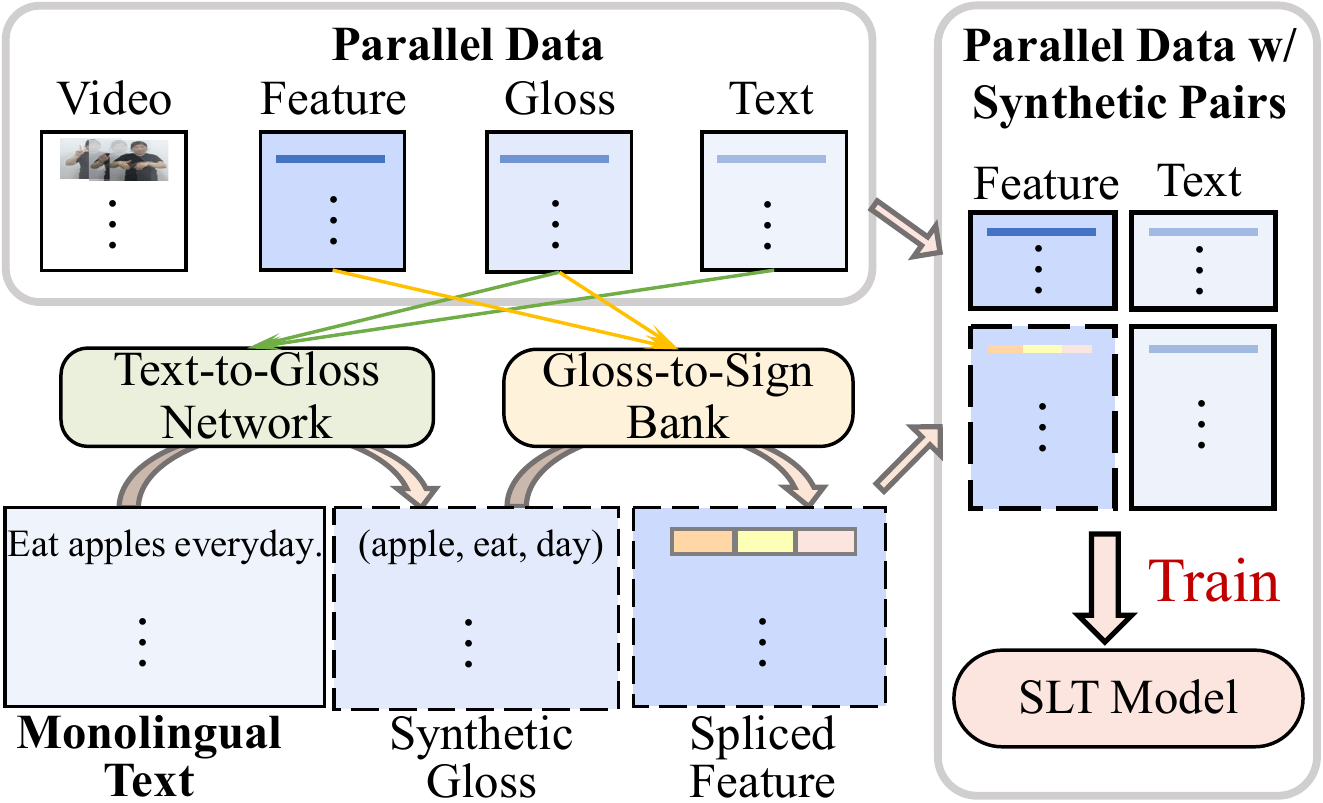}
   \caption{Illustration of the sign back-translation process. 
   }\label{fig:SignBT}
\vspace{-8pt}
\end{figure}

\textbf{Sign Back-Translation.}
Given an SLT corpus, parallel pairs of sign videos $\mathcal{X}$ and spoken language texts $\mathcal{Y}$ 
are converted to $(\mathcal{F}, \mathcal{Y})$ pairs through a sign embedding network. 
Meanwhile, monolingual spoken language texts $\mathcal{Y'}$ are collected, sharing a similar vocabulary with $\mathcal{Y}$. 
The following target is to generate synthetic pairs $(\mathcal{F'}_\text{syn}, \mathcal{Y'})$ with monolingual data $\mathcal{Y'}$, as depicted in Figure~\ref{fig:SignBT}. 

First, we train a text-to-gloss (T2G) network with existing parallel pairs of $(\mathcal{Y}, \mathcal{G})$ for back-translation. 
Then, the collected spoken language texts $\mathcal{Y'}$ are first translated to sign gloss texts $\mathcal{G'}_\text{syn}$. 
We splice gloss pieces from the G2S bank into sign embedding sequences $\mathcal{F'}_\text{syn}$ according to $\mathcal{G'}_\text{syn}$. 
As each gloss may have multiple feature pieces in G2S bank, we randomly sample one piece from them for splicing. 
In different training epochs, the spliced feature sequences of the same synthetic gloss sequence are different due to the random selections.
It largely enriches the feature combinations in the source domain. 

Finally, we mix synthetic pairs $(\mathcal{F'}_\text{syn}, \mathcal{Y'})$ with annotated pairs $(\mathcal{F}, \mathcal{Y})$ together for SLT training. Notably, the texts in the decoder side always comes from a real corpus.

\section{The Proposed CSL-Daily Dataset} \label{sec:CSL-Daily}

\begin{table}[tp]
   \centering
   \footnotesize
   \caption{Key statistics of the CSL-Daily split. 
   (OOV: out-of-vocabulary, \eg, words that occur in Dev set but not in Train set. Singleton: words that only occur once in Train unique sentences.)} \label{tab:csl-daily}
   \setlength{\tabcolsep}{2pt}{
      \begin{tabular}{l|rrr|rrr}
      \hline
      \multirow{2}{*}{}   & \multicolumn{3}{c|}{Sign Gloss} & \multicolumn{3}{c}{Chinese}  \\ 
                          & Train    & Dev   & Test  &  Train  & Dev  & Test                        \\  \hline
      segments            & 18,401   &1,077  &1,176  &  \multicolumn{3}{c}{$\longleftarrow$ same}        \\ 
      duration (h)        & 20.62    & 1.24  & 1.41  &  \multicolumn{3}{c}{$\longleftarrow$ same}        \\
      frames              &2,227,178 &134,530&153,074&  \multicolumn{3}{c}{$\longleftarrow$ same}        \\  \hline  
      vocab. size         &  2,000   &1,344  &1,345  &  2,343 &  1,358  &  1,358                         \\ 
      total words/chars   &133,714   &8,173  &9,002  & 291,048& 17,304  & 19,288                          \\ 
      total OOVs          &  -     &  0    &  0    &  -   &   64    &    69                          \\  
      unique sentences    &  6,598   &  797  &  798  &  6,598   &  797  &  798       \\ 
      singletons          &  247     &  -    &  -    &  418   &   -     &    -                          \\ 
      \hline
   \end{tabular}
  }
\vspace{-8pt}
\end{table}

CSL-Daily aims to offer the community a new large-scale sign language corpus, 
which is appropriate for both practical application and academic research. 
In this section, the details of dataset production are elaborated. 

\subsection{Data Collection}

The content of our corpus mainly revolves around the daily life of the deaf community. 
It covers a wide range of topics, including family life, medical care, school life, bank service, shopping, social contact and so on. 
 
Deaf community involvement is essential in developing sign language corpus~\cite{bragg2019slp}.
We invite a professional team composed of an expert in the field of sign language linguistics and several sign language teachers to help design the specific content and produce reference texts for recording guidance. 
The texts are mainly collected from some Chinese Sign Language textbooks and test material, and partly from some Chinese corpora. 

There are 10 signers participating in the video recording work. 
They are all native signers from the deaf community and 4 of them are engaged in sign language education. 
To remove the ambiguity of meanings, sign videos of one senior signer are recorded in advance as reference. 
After watching guidance videos, each reference text is signed again by one or two signers. 
No signers sign the same reference text twice. 
The requirement for signers is to ensure the natural expression of sign language and describe the content in reference texts as fully as possible. 

The resolution of recorded videos is $1920\!\times\!1080$ and the frame rate is 30 FPS. 
The motionless frames in the beginning and the end of a video are cut off carefully. 

\subsection{Annotation}

Our CSL-Daily provides two levels of annotation, \ie, sign gloss and spoken language translation.
Our annotation work relies on the cooperation of senior native signers and authors of this work. 
First, each sign running in videos is annotated with a Chinese word which has the similar meaning. 
Then, we adopt two strategies to merge the sign gloss with the same visual expression. 
One is to check the glosses with the similar meaning. 
The other is to train and test a sign-to-gloss network on the dataset. 
With the confusion matrix of predicted gloss, we focus the top-k confused pairs and check if they share the same sign indeed. 
With three rounds of double-checking, we reduce the vocabulary size of annotated sign glosses from $>\!3$k to $2$k. 
Then, spoken language translation annotation is conducted according to original reference texts and sign gloss annotations.

The detailed statistics of the dataset is shown in Table~\ref{tab:csl-daily}. 
In addition, a sign dictionary (SignDict) is produced. 
Each non-single sign is recorded by 4 sign teachers.  
The SignDict can be used for tasks like sign spotting, sign segmentation, isolated SLR and gloss-free SLT in the future. 
It can also serve as a reference collection for qualitative analysis of continuous sign language related tasks.

\section{Experiments} \label{sec:experiment}

\subsection{Experimental Setup}

\begin{table}[tp]
   \centering
   \footnotesize
   \caption{Statistics of the training data. (OOV-\%: the ratio of words or characters which are out of the parallel data vocabulary.)} \label{tab:trainingdata}
   \setlength{\tabcolsep}{3pt}{
      \begin{tabular}{c|rcl}
      \hline
                                       & Amount  & OOV (\%)  & Source                     \\  \hline 
      DGS$\leftrightarrow$ German      & 7,096   &  -        & PHOENIX-2014T  \\ 
      German texts                     & 212,247 & 7.07\%    & Wiki, weather forecast website  \\ \hline \hline
      CSL$\leftrightarrow$Chinese      & 18,402  &  -        & CSL-Daily             \\ 
      Chinese texts                    & 566,682 & 1.80\%    & Wiki, WebText in CLUE~\cite{xu-etal-2020-clue}             \\ 
      \hline
   \end{tabular}
  }
\vspace{-8pt}
\end{table}

\textbf{Dataset.} 
We mainly conduct ablation studies and evaluate our method on CSL-Daily. 
Experimental analysis on PHOENIX-2014T~\cite{slt-nslt-cihan18} is also reported. 
PHOENIX-2014T is a large-scale SLT corpus composed of German Sign Language (Deutsche Gebärdensprache, DGS). 
It is an extended version of PHOENIX-2014~\cite{phoenix2014} and contains parallel sign videos, gloss annotations and their German translations. 
The split of videos for Train, Dev and Test is 7096, 519 and 642, respectively. 
The vocabulary size is 1115 for sign gloss and 3000 for German. 

\textbf{Training Data.} 
Except the sentences in datasets, 
the majority are from the open wikipedia corpus~\cite{WMT19findings} (see Table~\ref{tab:trainingdata}). 
To be close to the datasets' topic, we also collect some texts from a German weather forecast website and extract a subset about trivia in daily lives from CLUE corpus~\cite{xu-etal-2020-clue}. 

\textbf{Evaluation.}
To assess the sign embedding layer, we adopt Word Error Rate (WER) as the metric of measuring the similarity between the predicted gloss sequence and the ground truth. 
To measure the SLT performance, we select BLEU~\cite{papineni2002bleu} and ROUGE~\cite{lin-2004-rouge} scores, commonly used in NMT. Here, BLEU is calculated with n-grams from 1 to 4. ROUGE refers to ROUGE-L F1-Score~\cite{lin-2004-rouge}.

\textbf{Sub-Problem Definition.} 
In this paper, we mainly discuss two sub-problems of SLT as follows, 
\begin{enumerate}[itemsep=0pt,topsep=2pt,parsep=1pt]
   \item Sign-to-Text (S2T): It predicts the spoken language translations directly from sign embedding sequences in an end-to-end pipeline.  
   \item Sign-to-Gloss-to-Text (S2G2T): 
               It resorts to sign gloss as an intermediate state. 
               The G2T network is trained with sign glosses predicted from an S2G network. 
\end{enumerate}

\subsection{Implementation Details}

\begin{table}[tp] 
   \centering
   \footnotesize
   \caption{Temporal Inception Network Architecture~\cite{cui-tmm19} (TIN) for sign language embedding. 
            1D Batch Norm (BN) layer is added after each temporal convolution layer.} \label{tab:arch}
  \setlength{\tabcolsep}{4.5pt}{
   \begin{tabular}{c|cc|c}
     \hline
      Layer              &  Stride    & Kernel   &   Output Size   \\  \hline \hline
      Input              &    -       &   -      &  $T\times224\times224\times3$ \\ 
      Inception Blocks w/ BN   &  $1,32,32$ & -        &  $T\times7\times7\times1024$ \\ 
      Global AvgPooling2D&  $1, 7, 7$ & $1,7,7$  &  $T\times1024$ \\ 
      Conv1D-BN1D-ReLU   &  $1,1,1$   & $5,1,1$  &  $T\times512$ \\ 
      MaxPooling1D       &  $2,1,1$   & $2,1,1$  &  $(T/2)\times512$ \\ 
      Conv1D-BN1D-ReLU   &  $1,1,1$   & $5,1,1$  &  $(T/2)\times512$ \\ 
      MaxPooling1D       &  $2,1,1$   & $2,1,1$  &  $(T/4)\times512$ \\ \hline \hline
      Transformer Encoder&    -       &  -       &  $(T/4)\times512$ \\  
      Fully Connection   &    -       &  -       &  $(T/4)\times C$ \\  \hline 
     \end{tabular}
  }
\end{table}

\begin{figure}[tp]
   \centering
   \includegraphics[width=0.22\textwidth]{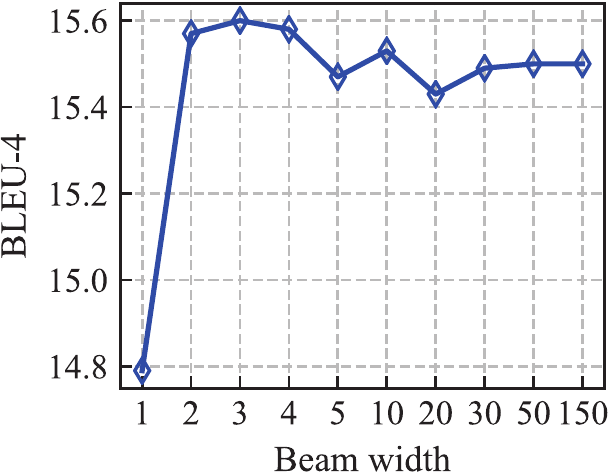}
   \includegraphics[width=0.23\textwidth]{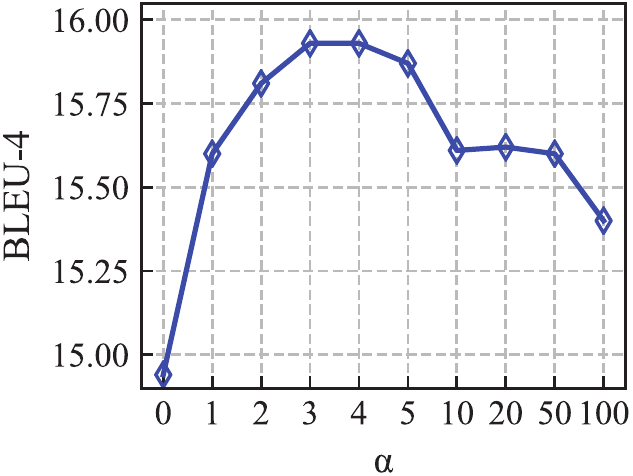}
   \caption{The effect of beam width and length penalty $\alpha$ on CSL-Daily Dev set under S2G2T setting.}\label{fig:alphabeam}
\vspace{-8pt}
\end{figure}

\textbf{Sign Embedding Layer.} 
The input frames are resized to $224\times224$. 
For data augmentation, we use random shift and random discard or copy of $20\%$ frames. 
The architecture of our sign embedding layer is presented in Table~\ref{tab:arch}. 
The pre-trained weights on ImageNet are loaded for initialization. 
The encoder with a classifier is only for pre-training and will be discarded in the following SLT experiments.    

\textbf{Transformer.}
In our experiments, the setting of all transformer layers is the same. 
The hidden size is 512 and the feed-forward size is 2048. 
Each layer has 8 attention heads which is the basic setting of transformer~\cite{vaswani2017attention}. 
The dropout rate is all set to 0.1 to alleviate over-fitting. 

\textbf{Optimization.}
The sign embedding layer is trained end-to-end under CTC Loss with batch size 2. 
No iterative training~\cite{DPD_icme19}, online refining~\cite{cslr-fcn-20fully} or temporal sampling~\cite{cslr-20-Stochastic} are used. 
We use Adam optimizer~\cite{adam} and set the weight decay to $1\times10^{-6}$. 
The learning rate is initialized as $5\times10^{-5}$. 
It will be reduced by a factor of 0.5 until $2\times10^{-6}$ when WER of Dev stops decreasing for 3 epochs. 
Experiments are run on 4 Titan RTX GPUs. 

The transformer is trained end-to-end under masked cross-entropy loss~\cite{trans-vaswani17attention} with batch size 32. 
The rate of label smoothing~\cite{labelsmooth,trans-vaswani17attention} is 0.1.  
We use Adam optimizer with no weight decay. 
The learning rate is fixed to $5\times10^{-5}$. 
Experiments are run on 1 Titan RTX GPU. 

\textbf{Inference.} 
For decoding in the inference process, we use the beam search strategy~\cite{googletranslation}. It is combined with a length penalty $\alpha$~\cite{googletranslation} for length normalization. 
For PHOENIX-2014T, we set beam width to 3 and $\alpha$ to 1, following~\cite{slt-trans-cihan20}. 
In contrast, Chinese sentences are longer in character-level tokenization. 
We search the combinations in Figure~\ref{fig:alphabeam} and use beam width of 3 and length penalty $\alpha$ of 3.

\subsection{Ablation Study}
The ablation experiments are mainly conducted on CSL-Daily-Dev, presenting the characteristics of this new corpus.

\begin{table}[tp]
   \centering
   \footnotesize
   \caption{Evaluation of the S2G network combinations on WER (the lower the better). } \label{tab:S2G}
  \setlength{\tabcolsep}{6pt}{
   \begin{tabular}{cc|cr|cc}
     \hline
       \multicolumn{2}{c|}{S2G combination} & \multicolumn{2}{c|}{PH2014T} & \multicolumn{2}{c}{CSL-Daily} \\
     Sign Embedding &  Encoder    & Dev    & Test  & Dev   & Test     \\ \hline
     I3D            & Transformer & 32.6   & 33.2  & 45.4  & 44.3        \\                        
     TIN            & Transformer & 26.2   & 27.5  & 36.1  & 35.7    \\
     BN-TIN         & Transformer & 23.0   & 24.1  & 33.6  & 33.1   \\
     BN-TIN         & Conv1D      & 24.7   & 25.1  & 33.4  & 33.3 \\ 
     BN-TIN         & Bi-GRU      & \tb{22.7} & \tb{23.9}  & \tb{33.2}  & \tb{32.2} \\ \hline
     \end{tabular}
  }
\end{table}

\begin{table}[tp]
   \centering
   \footnotesize
   \caption{Performance of CSLR baselines on CSL-Daily. $*$ denotes that the results are based on our implementation. } \label{tab:cslr-csl}
   \setlength{\tabcolsep}{6pt}{

     \begin{tabular}{l|cc|cc}
     \hline
     \multirow{2}{*}{Method}                   & \multicolumn{2}{c|}{Dev} & \multicolumn{2}{c}{Test} \\
                                               & del/ins  & WER  &del/ins & WER   \\ \hline
     SubUNets~\cite{subunet}               & 14.8/3.0 & 41.4  & 14.6/2.8 & 41.0      \\
     LS-HAN$^*$~\cite{huang18han}              & 14.6/5.7 & 39.0  & 14.8/5.0 & 39.4    \\
     TIN-Iterative$^*$~\cite{cui-tmm19}        & 12.8/3.3 & \textbf{32.8}  & 12.5/2.7 & 32.4     \\
     Joint-SLRT~\cite{slt-trans-cihan20}   &10.3/4.4& 33.1 & 9.6/4.1 &  \textbf{32.0}   \\
     FCN-GFE$^*$~\cite{cslr-fcn-20fully}       &12.8/4.0 & 33.2 &12.6/3.7 & 32.5     \\ 
     BN-TIN+Transf. Encoder                          &13.9/3.4 & 33.6 &13.5/3.0 & 33.1   \\ \hline
     \end{tabular}
  }
\vspace{-8pt}
\end{table}

\textbf{Sign Language Embedding.} 
In Table~\ref{tab:S2G},  we investigate which kind of spatiotemporal combinations is suitable for sign embedding. 
The I3D model~\cite{I3D_Kinetics} achieves good performance in the action recognition task.  
However, with less spatial details, it still has a performance gap to 2D-CNN based methods. 
Unlike previous re-fining methods~\cite{cslr-fcn-20fully,cui-tmm19,DPD_icme19}, we use 1D batch normalization (BN) to mitigate the unstable activation in temporal structures. 
It achieves favorable performance under end-to-end training without bells and whistles. 
Hence, we adopt BN-TIN as our sign embedding layer. 
In Table~\ref{tab:cslr-csl}, we also provides some results of CSLR methods on CSL-Daily for reference. 

\textbf{Encoder-Decoder Framework.} 
In Table~\ref{tab:encoderdecoder}, we evaluate encoder-decoder networks with different architectures. 
For the sophisticated design of self-attention~\cite{trans-vaswani17attention}, the transformer-based SLT model achieves obvious advantage over previous recurrent neural network-based methods~\cite{seq2seq-attn-Bahdanau,luong-seq2seq}. 
We set the transformer-based network as our baseline model for the following experiments.

\begin{table}[tp]
   \centering
   \footnotesize
   \caption{Evaluation of different encoder-decoder frameworks on CSL-Daily. 
   (R: ROUGE, B-n: BLEU-n, the higher the better.)} \label{tab:encoderdecoder}
  \setlength{\tabcolsep}{5.5pt}{
   \begin{tabular}{l|ccccc}
     \hline
     S2G2T             & R      & B-1   & B-2   & B-3    & B-4     \\ \hline
     seq2seq w/ Bahdanau~\cite{seq2seq-attn-Bahdanau}  & 39.63  & 41.58 & 25.34 & 16.08  & 10.63    \\
     seq2seq w/ Luong~\cite{luong-seq2seq}     & 40.18  & 41.46 & 25.71 & 16.57  & 11.06      \\                        
     Transformer~\cite{trans-vaswani17attention}       & \tb{44.21}  & \tb{46.61} & \tb{32.11} & \tb{22.44}  & \tb{15.93}    \\
     \hline \hline
     S2T               & R      & B-1   & B-2   & B-3    & B-4     \\ \hline
     seq2seq w/ Bahdanau~\cite{seq2seq-attn-Bahdanau}  & 33.83  & 33.99 & 19.48 & 11.66  & 7.11   \\
     seq2seq w/ Luong~\cite{luong-seq2seq}     & 34.28  & 34.22 & 19.72 & 12.24  & 7.96      \\                        
     Transformer~\cite{trans-vaswani17attention}       & \tb{37.29}  & \tb{40.66} & \tb{26.56} & \tb{18.06}  & \tb{12.73}    \\
     \hline
     \end{tabular}
  }
\end{table}

\begin{table}[tp]
   \centering
   \footnotesize
   \caption{The number of epochs for warm-up on CSL-Daily. (Warm-up: mix all synthetic data with parallel data for training)} \label{tab:warmup}
  \setlength{\tabcolsep}{3pt}{
   \begin{tabular}{r|cccc|cccc}
     \hline
     warm-up  & \multicolumn{4}{c|}{S2G2T} & \multicolumn{4}{c}{S2T} \\ 
     \#epochs    & R      &  B-2 & B-3   & B-4   & R     & B-2 & B-3   & B-4    \\ \hline
     0 (0.4h)  & 44.21  & 32.11  & 22.44 & 15.93 & 37.29 & 26.56 & 18.06 & 12.73  \\
     1  (0.6h)       & 46.22  & 34.47  & 24.70 & 18.06 & 42.69 & 31.72 & 22.03 & 15.64  \\
     5  (1.6h)       & 47.68  & 35.51  & 25.58 & 18.73 & 46.56 & 34.33 & 24.54 & 17.98  \\ 
     10 (2.9h)       & 47.88  & 36.08  & 26.20 & 19.38 & 47.75 & 35.17 & 25.58 & 19.11  \\
     20 (5.4h)       & 48.01  & 35.57  & 25.82 & 19.18 & 48.55 & 36.07 & 26.24 & 19.61  \\                        
     \tb{50} (12.9h)  & \tb{48.38}  & \tb{36.16}  & \tb{26.26} & \tb{19.53} & 48.77 & 36.63 & 26.90 & 20.20  \\
     100 (25.4h) & 47.83  & 36.05  & 26.17 & 19.42 & \tb{49.09} & \tb{36.91} & \tb{27.20} & \tb{20.50}  \\
     \hline
     \end{tabular}
  }
\end{table}

\begin{table}[tp]
   \centering
   \footnotesize
   \caption{The ratio of synthetic data to parallel data for the training process after warm-up. } \label{tab:ratio}
  \setlength{\tabcolsep}{3pt}{
      \begin{tabular}{l|cccc|cccc}
         \hline
                  & \multicolumn{4}{c|}{S2G2T} & \multicolumn{4}{c}{S2T} \\ 
         ratio    & R      &  B-2 & B-3   & B-4   & R     & B-2 & B-3   & B-4    \\ \hline                       
         0.0: 1  & \tb{48.38}  & \tb{36.16}  & \tb{26.26} & \tb{19.53} & 48.77 & 36.63 & 26.90 & 20.20  \\
         0.1: 1  & 46.97  & 35.49  & 25.80 & 19.20 & \tb{49.49}  & \tb{37.23}  & \tb{27.51} & \tb{20.80}  \\
         0.5: 1  & 46.30  & 34.77  & 25.19 & 18.82 & 49.15  & 36.88  & 27.23 & 20.64  \\
         1.0: 1  & 45.76  & 34.43  & 24.65 & 18.22 & 49.21  & 36.38  & 26.80 & 20.27  \\
         \hline
     \end{tabular}
  }
  \vspace{-4pt}
\end{table}

\textbf{The Participation of Synthetic Data.} 
We generate synthetic pairs from texts in Table~\ref{tab:trainingdata}. 
They are over 30 times the amount of annotated parallel data. 
If directly mixing them for training with no adjustment, the noise in synthetic pairs will largely disturb the model learning. 
Hence, we first use all data for warm-up and then train models until convergence with less synthetic pairs. 

\begin{figure}[tp]
   \centering
   \includegraphics[width=0.218\textwidth]{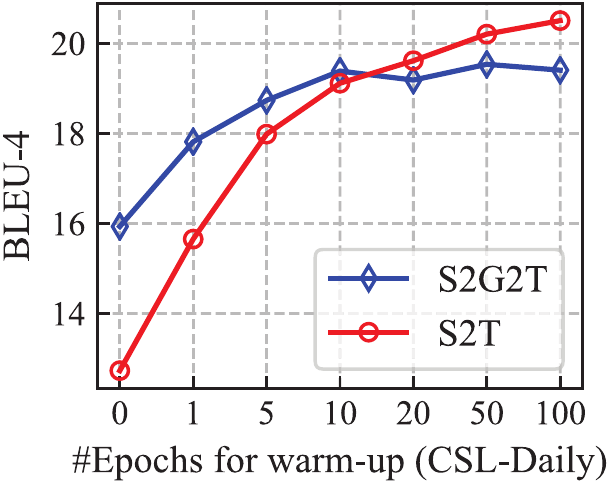}
   \includegraphics[width=0.24\textwidth]{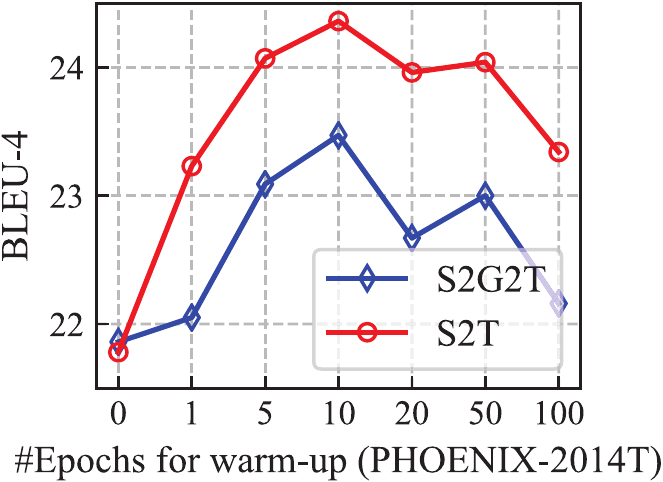}
   \caption{The effect of warm-up on different datasets.}\label{fig:warmup}
\end{figure}

In Table~\ref{tab:warmup}, we evaluate the SLT performance with different warm-up epochs on CSL-Daily. 
Even with only one warm-up epoch, the performance gain brought by synthetic data is obvious across all metrics. 
With the increasing of warm-up epochs, the final performance improves gradually. 
To verify the universality, the effect of warm-up on different datasets is presented in Figure~\ref{fig:warmup}. 
Unlike on CSL-Daily, large warm-up epochs do not bring further improvement on PHOENIX-2014T but a slight decrease in the BLEU score. 
Considering that the topic of PHONIEX-2014T is all around weather forecast, 
it may constrain the learning of linguistic from synthetic data. 
Although we do collect some sentences about the weather, they account for a small slice of all data. 
Considering training time, we use 50 warm-up epochs for CSL-Daily and 10 for PHOENIX-2014T. 

After warm-up, we use a small portion of synthetic data for training, which is sampled randomly after each epoch. 
In Table~\ref{tab:ratio}, we evaluate several mix ratios of training data. 
When synthetic data account for a small ratio, the S2T models get better performance, compared to models trained with directly giving them up. 
In contrast, the participation of synthetic data after warm-up consistently harms the S2G2T model. 
The noise comes mainly from the synthetic part.  
We argue that the noise in sparse gloss-level is difficult for the model to handle, while the noise in dense feature-level enables better generalization instead. 

\begin{table}[tp]
   \centering
   \footnotesize
   \caption{The quantity of synthetic data compared to parallel data participating in the training process on CSL-Daily.} \label{tab:quantity}
  \setlength{\tabcolsep}{3pt}{
   \begin{tabular}{r|cccc|cccc}
     \hline
                 & \multicolumn{4}{c|}{S2G2T} & \multicolumn{4}{c}{S2T} \\ 
     quantity    & R      &  B-2 & B-3   & B-4   & R     & B-2 & B-3   & B-4    \\ \hline
     0$\times$   & 44.21  & 32.11  & 22.44 & 15.93 & 37.29 & 26.56 & 18.06 & 12.73  \\
     1$\times$   & 45.62  & 33.84  & 23.98 & 17.30 & 40.66 & 29.97 & 21.03 & 15.24  \\
     5$\times$   & 46.57  & 34.85  & 24.88 & 18.22 & 45.47 & 33.86 & 24.39 & 18.05  \\ 
     10$\times$  & 47.13  & 35.42  & 25.28 & 18.50 & 46.85 & 35.08 & 25.80 & 19.43  \\
   \textgreater30$\times$  & \tb{48.38}  & \tb{36.16}  & \tb{26.26} & \tb{19.53} & \tb{49.49}  & \tb{37.23}  & \tb{27.51} & \tb{20.80}  \\
   \hline
     \end{tabular}
  }
\end{table}

\begin{table}[tp]
   \centering
   \footnotesize
   \caption{The quality of synthetic data on CSL-Daily. The number in ($\cdot$) denotes the BLEU-4 score of T2G networks for SignBT.} \label{tab:qualtiy}
   \setlength{\tabcolsep}{2pt}{
   \begin{tabular}{l|cccc|cccc}
     \hline
                    & \multicolumn{4}{c|}{S2G2T} & \multicolumn{4}{c}{S2T} \\ 
                        & R      &  B-2 & B-3   & B-4     & R     & B-2 & B-3   & B-4    \\ \hline
     w/o synthetic  & 44.21  & 32.11  & 22.44 & 15.93 & 37.29 & 26.56 & 18.06 & 12.73  \\
     blank input        & 45.83  & 33.49  & 23.99 & 17.36 & 41.22 & 30.44 & 21.60 & 15.77  \\
     low (3.05)     & 46.31  & 34.41  & 24.78 & 18.21 & 43.78 & 31.76 & 22.85 & 16.91  \\
     medium (7.02)  & 47.64  & 35.56  & 25.77 & 19.08 & 46.15 & 33.96 & 24.66 & 18.50  \\ 
     High (11.63) & \tb{48.38}  & \tb{36.16}  & \tb{26.26} & \tb{19.53} & \tb{49.49}  & \tb{37.23}  & \tb{27.51} & \tb{20.80}  \\
     \hline
     \end{tabular}
  }
  \vspace{-4pt}
\end{table}

\begin{table*}[tp]
   \centering
   \footnotesize
   \caption{Comparison with methods for SLT on PHOENIX-2014T (the higher the better).} \label{tab:slt-ph}
   \setlength{\tabcolsep}{4pt}{
      \begin{tabular}{l|ccccc|ccccc}
   \hline
   & \multicolumn{5}{c|}{Dev} & \multicolumn{5}{c}{Test} \\
   S2G2T                & ROUGE  & BLEU-1  & BLEU-2 & BLEU-3 & BLEU-4 & ROUGE  & BLEU-1  & BLEU-2 & BLEU-3 & BLEU-4\\   \hline 
   SL-Luong~\cite{slt-nslt-cihan18}                    & 44.14 & 42.88 & 30.30 & 23.02 & 18.40 & 43.80 & 43.29 & 30.39 & 22.82 & 18.13 \\
   SL-Transf.~\cite{slt-trans-cihan20}                 & -     & 47.73 & 34.82 & 27.11 & 22.11 & -     & 48.47 & 35.35 & 27.57 & 22.45 \\ 
   BN-TIN-Transf.$^2$ (baseline)                        & 47.83 & 47.72 & 34.78 & 26.94 & 21.86 & 47.98 & 47.74 & 35.27 & 27.59 & 22.54 \\
   \textbf{BN-TIN-Transf.$^2$+BT} (ours)                & \tb{49.53} & \tb{49.33} & \tb{36.43} & \tb{28.66} & \tb{23.51} & \tb{49.35} & \tb{48.55} & \tb{36.13} & \tb{28.47} & \tb{23.51} \\  \hline \hline
   S2T                & ROUGE  & BLEU-1  & BLEU-2 & BLEU-3 & BLEU-4 & ROUGE  & BLEU-1  & BLEU-2 & BLEU-3 & BLEU-4\\   \hline 
   SL-Luong~\cite{slt-nslt-cihan18}                    & 31.80 & 31.87 & 19.11 & 13.16 & 9.94  & 31.80 & 32.24 & 19.03 & 12.83 & 9.58 \\  
   Joint-SLRT~\cite{slt-trans-cihan20}                 & -     & 47.26 & 34.40 & 27.05 & 22.38 & -     & 46.61 & 33.73 & 26.19 & 21.32 \\ 
   TSPNet-Joint~\cite{tspnet-nips20}                   & -     &   -   &  -    &   -   &   -   & 34.96 & 36.10 & 23.12 & 16.88 & 13.41 \\  
   BN-TIN-Transf. (baseline)                            & 46.87 & 46.90 & 33.98 & 26.49 & 21.78 & 46.98 & 47.57 & 34.64 & 26.78 & 21.68 \\
   \tb{BN-TIN-Transf.+SignBT} (ours)                    & \tb{50.29} & \tb{51.11} & \tb{37.90} & \tb{29.80} & \tb{24.45} & \tb{49.54} & \tb{50.80} & \tb{37.75} & \tb{29.72} & \tb{24.32} \\ \hline
   MCT~\cite{slt-camgoz2020multi}                    & 45.90 & -  & - & - & 19.51 & 43.57 & - & - & - & 18.51 \\ 
   STMC-T~\cite{stmc-tmm}                              & 48.24 & 47.60 & 36.43 & 29.18 & 24.09 & 46.65 & 46.98 & 36.09 & 28.70 & 23.65 \\ 
   \hline
   \end{tabular}
   }
\end{table*}

\begin{table*}[tp]
   \centering
   \footnotesize
   \caption{Comparison with methods for SLT on CSL-Daily (the higher the better).} \label{tab:slt-csl}
   \setlength{\tabcolsep}{4pt}{
      \begin{tabular}{l|ccccc|ccccc}
   \hline
         & \multicolumn{5}{c|}{Dev} & \multicolumn{5}{c}{Test} \\
   S2G2T                & ROUGE  & BLEU-1  & BLEU-2 & BLEU-3 & BLEU-4 & ROUGE  & BLEU-1  & BLEU-2 & BLEU-3 & BLEU-4\\   \hline 
   SL-Luong~\cite{slt-nslt-cihan18}                    & 40.18 & 41.46 & 25.71 & 16.57 & 11.06 & 40.05 & 41.55 & 25.73 & 16.54 & 11.03 \\  
   SL-Transf.~\cite{slt-trans-cihan20}                 & 44.18 & 46.82 & 32.22 & 22.49 & 15.94 & 44.81 & 47.09 & 32.49 & 22.61 & 16.24 \\ 
   BN-TIN-Transf.$^2$ (baseline)                        & 44.21 & 46.61 & 32.11 & 22.44 & 15.93 & 44.78 & 46.85 & 32.37 & 22.57 & 16.25 \\
   \tb{BN-TIN-Transf.$^2$+BT} (ours)                        & \tb{48.38} & \tb{50.97} & \tb{36.16} & \tb{26.26} & \tb{19.53} & \tb{48.21} & \tb{50.68} & \tb{36.00} & \tb{26.20} & \tb{19.67} \\ 
   \hline \hline
   S2T   & ROUGE  & BLEU-1  & BLEU-2 & BLEU-3 & BLEU-4 & ROUGE  & BLEU-1  & BLEU-2 & BLEU-3 & BLEU-4\\   \hline 
   SL-Luong~\cite{slt-nslt-cihan18}                    & 34.28 & 34.22 & 19.72 & 12.24 &  7.96 & 34.54 & 34.16 & 19.57 & 11.84 &  7.56 \\
   Joint-SLRT~\cite{slt-trans-cihan20}                 & 37.06 & 37.47 & 24.67 & 16.86 & 11.88 & 36.74 & 37.38 & 24.36 & 16.55 & 11.79 \\ 
   BN-TIN-Transf. (baseline)                            & 37.29 & 40.66 & 26.56 & 18.06 & 12.73 & 37.67 & 40.74 & 26.96 & 18.48 & 13.19 \\
   \tb{BN-TIN-Transf.+SignBT} (ours)                    & \tb{49.49} & \tb{51.46} & \tb{37.23} & \tb{27.51} & \tb{20.80} & \tb{49.31} & \tb{51.42} & \tb{37.26} & \tb{27.76} & \tb{21.34} \\ 
   \hline
   \end{tabular}
   }
\end{table*}

\textbf{The Quantity and Quality of Synthetic Data.} 
In Table~\ref{tab:quantity}, we train the SLT network with different quantity of synthetic data. 
The performance improves steadily when increasing synthetic data volume.

Besides, we analyse the performance variations using different quality of synthetic data. 
We train three text-to-gloss (T2G) networks with different epochs, \emph{i.e.,} low, medium and high, whose BLEU-4 scores are 3.05, 7.02, and 11.63, respectively. 
Those three T2G networks are then used to generate synthetic data of different qualities. 
As shown in Table~\ref{tab:qualtiy}, the metric scores of SLT model are higher with higher quality synthetic data.
We also simulate the worst condition by pairing monolingual texts with blank input for training (corresponding to ``blank input" in Table~\ref{tab:qualtiy}). 
While achieving a small gain, it has a performance gap compared to the models trained with synthetic data. 
It verifies that the synthetic pairs from our SignBT do make effects on the seq2seq learning, rather than simple enhancement in language modelling.

\subsection{Comparison with State-of-the-art Methods}
Our SignBT mechanism is dedicated to the S2T setting, which directly translates spoken language from videos. 
The results of back-translation on S2G2T are also provided. 

\textbf{Evaluation on PHOENIX-2014T:}
In Table~\ref{tab:slt-ph}, we compare our approach with SLT methods on PHOENIX-2014T. 
MCT~\cite{slt-camgoz2020multi} and STMC-T~\cite{stmc-tmm} are evaluated under multi-cue setting. 
TSPNet-Joint~\cite{tspnet-nips20} explores gloss-free S2T methods with word-level sign language corpus for pre-training. 
Joint-SLRT~\cite{slt-trans-cihan20} jointly models CSLR and SLT problems in one framework.
Our baseline model is at the same level as previous methods. 
The SignBT approach gives an improvement of 2.6 BLEU-4 points on both sets.  

\textbf{Evaluation on CSL-Daily:}
In Table~\ref{tab:slt-csl}, we compare our approach with SLT methods on CSL-Daily. 
The performance boost with SignBT on CSL-Daily is more significant than that on PHOENIX-2014T, which is attributed to two factors. 
On one hand, Chinese sentences are built on three levels, \emph{i.e.,} character, word, and sentence.  
The number of unique characters in sentences is 2.3K, but they have over 8K combinations in word-level. 
On the other hand, the vocabulary size of sign words in videos also exceeds 2K. 
The vocabulary sizes of both sides are quite large. 
When our SignBT approach serves as a data-augmentation method for the encoder-decoder framework,
it is more effective when dealing with the large-scale vocabulary problem.

\section{Conclusion} \label{sec:conclusion}

In this paper, we propose to improve the translation quality with monolingual data, 
which is rarely investigated in SLT. 
By designing a SignBT pipeline, we convert massive spoken language texts into source sign sequences. 
The synthetic pairs are treated as additional training data to alleviate the shortage of parallel data in training. 
With no change on the network architectures, our approach can be easily applied to encoder-decoder based SLT methods. 
Moreover, we contribute a large-scale SLT dataset with diverse topics and complete annotations. 
Extensive experiments demonstrate the significance of our sign back-translation approach.

\footnotesize {\flushleft \bf Acknowledgements}. 
This work was supported in part by the National Natural Science Foundation of China under Contract U20A20183, 61836011 and 62021001, and in part by the Youth Innovation Promotion Association CAS under Grant 2018497. 
It was also supported by the GPU cluster built by MCC Lab of Information Science and Technology Institution, USTC.

{\small
\bibliographystyle{ieee_fullname}
\bibliography{reference}

\begin{thebibliography}{10}\itemsep=-1pt

\bibitem{bsl1k-20}
Samuel Albanie, G{\"u}l Varol, Liliane Momeni, Triantafyllos Afouras, Joon~Son
  Chung, Neil Fox, and Andrew Zisserman.
\newblock {BSL-1K}: Scaling up co-articulated sign language recognition using
  mouthing cues.
\newblock {\em ECCV}, 2020.

\bibitem{seq2seq-attn-Bahdanau}
Dzmitry Bahdanau, Kyunghyun Cho, and Yoshua Bengio.
\newblock Neural machine translation by jointly learning to align and
  translate.
\newblock In {\em ICLR}, 2014.

\bibitem{WMT19findings}
Lo{\"\i}c Barrault, Ond{\v{r}}ej Bojar, Marta~R Costa-Juss{\`a}, Christian
  Federmann, Mark Fishel, Yvette Graham, Barry Haddow, Matthias Huck, Philipp
  Koehn, Shervin Malmasi, et~al.
\newblock Findings of the 2019 conference on machine translation (wmt19).
\newblock In {\em WMT}, 2019.

\bibitem{bragg2019slp}
Danielle Bragg, Oscar Koller, Mary Bellard, Larwan Berke, Patrick Boudreault,
  Annelies Braffort, Naomi Caselli, Matt Huenerfauth, Hernisa Kacorri, Tessa
  Verhoef, Christian Vogler, and Meredith~Ringel Morris.
\newblock Sign language recognition, generation, and translation: An
  interdisciplinary perspective.
\newblock In {\em ACM ASSETS}, 2019.

\bibitem{slt-camgoz2020multi}
Necati~Cihan Camgoz, Oscar Koller, Simon Hadfield, and Richard Bowden.
\newblock Multi-channel transformers for multi-articulatory sign language
  translation.
\newblock In {\em ECCVW}, 2020.

\bibitem{I3D_Kinetics}
Joao Carreira and Andrew Zisserman.
\newblock Quo vadis, action recognition? a new model and the kinetics dataset.
\newblock In {\em CVPR}, 2017.

\bibitem{cslr-fcn-20fully}
Ka~Leong Cheng, Zhaoyang Yang, Qifeng Chen, and Yu-Wing Tai.
\newblock Fully convolutional networks for continuous sign language
  recognition.
\newblock {\em ECCV}, 2020.

\bibitem{chung2017lip}
Joon~Son Chung, Andrew Senior, Oriol Vinyals, and Andrew Zisserman.
\newblock Lip reading sentences in the wild.
\newblock In {\em CVPR}, 2017.

\bibitem{subunet}
Necati Cihan~Camgoz, Simon Hadfield, Oscar Koller, and Richard Bowden.
\newblock Subunets: end-to-end hand shape and continuous sign language
  recognition.
\newblock In {\em ICCV}, 2017.

\bibitem{slt-nslt-cihan18}
Necati Cihan~Camgoz, Simon Hadfield, Oscar Koller, Hermann Ney, and Richard
  Bowden.
\newblock Neural sign language translation.
\newblock In {\em CVPR}, 2018.

\bibitem{slt-trans-cihan20}
Necati Cihan~Camgoz, Oscar Koller, Simon Hadfield, and Richard Bowden.
\newblock Sign language transformers: Joint end-to-end sign language
  recognition and translation.
\newblock In {\em CVPR}, 2020.

\bibitem{SLG-TMM-THU-20}
Runpeng Cui, Zhong Cao, Weishen Pan, Changshui Zhang, and Jianqiang Wang.
\newblock Deep gesture video generation with learning on regions of interest.
\newblock {\em TMM}, 22(10):2551--2563, 2020.

\bibitem{staged}
Runpeng Cui, Hu Liu, and Changshui Zhang.
\newblock Recurrent convolutional neural networks for continuous sign language
  recognition by staged optimization.
\newblock In {\em CVPR}, 2017.

\bibitem{cui-tmm19}
Runpeng Cui, Hu Liu, and Changshui Zhang.
\newblock A deep neural framework for continuous sign language recognition by
  iterative training.
\newblock {\em IEEE TMM}, 21(7):1880--1891, 2019.

\bibitem{CTCLoss}
Alex Graves, Santiago Fern{\'a}ndez, Faustino Gomez, and J{\"u}rgen
  Schmidhuber.
\newblock Connectionist temporal classification: labelling unsegmented sequence
  data with recurrent neural networks.
\newblock In {\em ICML}, 2006.

\bibitem{gulcehre2015using}
Caglar Gulcehre, Orhan Firat, Kelvin Xu, Kyunghyun Cho, Loic Barrault, Huei-Chi
  Lin, Fethi Bougares, Holger Schwenk, and Yoshua Bengio.
\newblock On using monolingual corpora in neural machine translation.
\newblock {\em arXiv preprint arXiv:1503.03535}, 2015.

\bibitem{huang18han}
Jie Huang, Wengang Zhou, Qilin Zhang, Houqiang Li, and Weiping Li.
\newblock Video-based sign language recognition without temporal segmentation.
\newblock In {\em AAAI}, 2018.

\bibitem{hunt1996unit}
Andrew~J Hunt and Alan~W Black.
\newblock Unit selection in a concatenative speech synthesis system using a
  large speech database.
\newblock In {\em ICASSP}, 1996.

\bibitem{yukai3d}
Shuiwang Ji, Wei Xu, Ming Yang, and Kai Yu.
\newblock {3D} convolutional neural networks for human action recognition.
\newblock {\em IEEE TPAMI}, 35(1):221--231, 2013.

\bibitem{msasl-bmvc19}
Hamid Reza~Vaezi Joze and Oscar Koller.
\newblock Ms-asl: A large-scale data set and benchmark for understanding
  american sign language.
\newblock {\em BMVC}, 2019.

\bibitem{adam}
Diederik~P Kingma and Jimmy Ba.
\newblock Adam: A method for stochastic optimization.
\newblock In {\em ICLR}, 2015.

\bibitem{hmm-koller-tpami19}
Oscar Koller, Cihan Camgoz, Hermann Ney, and Richard Bowden.
\newblock Weakly supervised learning with multi-stream cnn-lstm-hmms to
  discover sequential parallelism in sign language videos.
\newblock {\em IEEE TPAMI}, 42(9):2306--2320, 2020.

\bibitem{phoenix2014}
Oscar Koller, Jens Forster, and Hermann Ney.
\newblock Continuous sign language recognition: Towards large vocabulary
  statistical recognition systems handling multiple signers.
\newblock {\em CVIU}, 141:108--125, 2015.

\bibitem{wlasl-asu20}
Dongxu Li, Cristian Rodriguez, Xin Yu, and Hongdong Li.
\newblock Word-level deep sign language recognition from video: A new
  large-scale dataset and methods comparison.
\newblock In {\em WACV}, 2020.

\bibitem{tspnet-nips20}
Dongxu Li, Chenchen Xu, Xin Yu, Kaihao Zhang, Ben Swift, Hanna Suominen, and
  Hongdong Li.
\newblock Tspnet: Hierarchical feature learning via temporal semantic pyramid
  for sign language translation.
\newblock In {\em NeurIPS}, 2020.

\bibitem{islr-transfer-cvpr2020}
Dongxu Li, Xin Yu, Chenchen Xu, Lars Petersson, and Hongdong Li.
\newblock Transferring cross-domain knowledge for video sign language
  recognition.
\newblock In {\em CVPR}, 2020.

\bibitem{lin-2004-rouge}
Chin-Yew Lin.
\newblock {ROUGE}: A package for automatic evaluation of summaries.
\newblock In {\em ACLW}, 2004.

\bibitem{luong-seq2seq}
Thang Luong, Hieu Pham, and Christopher~D. Manning.
\newblock Effective approaches to attention-based neural machine translation.
\newblock In {\em EMNLP}, 2015.

\bibitem{labelsmooth}
Rafael M{\"u}ller, Simon Kornblith, and Geoffrey~E Hinton.
\newblock When does label smoothing help?
\newblock In {\em NeurIPS}, 2019.

\bibitem{cslr-20-Stochastic}
Zhe Niu and B.~K. Mak.
\newblock Stochastic fine-grained labeling of multi-state sign glosses for
  continuous sign language recognition.
\newblock In {\em ECCV}, 2020.

\bibitem{ong2005automatic}
Sylvie~CW Ong and Surendra Ranganath.
\newblock Automatic sign language analysis: A survey and the future beyond
  lexical meaning.
\newblock {\em IEEE TPAMI}, 6:873--891, 2005.

\bibitem{papineni2002bleu}
Kishore Papineni, Salim Roukos, Todd Ward, and Wei-Jing Zhu.
\newblock Bleu: a method for automatic evaluation of machine translation.
\newblock In {\em ACL}, 2002.

\bibitem{pu2020boosting}
Junfu Pu, Wengang Zhou, Hezhen Hu, and Houqiang Li.
\newblock Boosting continuous sign language recognition via cross modality
  augmentation.
\newblock In {\em ACM MM}, 2020.

\bibitem{ian}
Junfu Pu, Wengang Zhou, and Houqiang Li.
\newblock Iterative alignment network for continuous sign language recognition.
\newblock In {\em CVPR}, 2019.

\bibitem{SLG-ECCV20}
Ben Saunders, Necati~Cihan Camgoz, and Richard Bowden.
\newblock Progressive transformers for end-to-end sign language production.
\newblock In {\em ECCV}, 2020.

\bibitem{backtranslationACL16}
Rico Sennrich, Barry Haddow, and Alexandra Birch.
\newblock Improving neural machine translation models with monolingual data.
\newblock In {\em ACL}, 2016.

\bibitem{fingerspelling19iccv}
Bowen Shi, Aurora Martinez~Del Rio, Jonathan Keane, Diane Brentari, Greg
  Shakhnarovich, and Karen Livescu.
\newblock Fingerspelling recognition in the wild with iterative visual
  attention.
\newblock In {\em ICCV}, 2019.

\bibitem{MonkeyNet-CVPR19}
Aliaksandr Siarohin, Stéphane Lathuilière, Sergey Tulyakov, Elisa Ricci, and
  Nicu Sebe.
\newblock Animating arbitrary objects via deep motion transfer.
\newblock In {\em CVPR}, 2019.

\bibitem{vgg-two-stream}
Karen Simonyan and Andrew Zisserman.
\newblock Two-stream convolutional networks for action recognition in videos.
\newblock In {\em Neurips}, 2014.

\bibitem{include-20-ISL}
Advaith Sridhar, Rohith~Gandhi Ganesan, Pratyush Kumar, and Mitesh Khapra.
\newblock Include: A large scale dataset for indian sign language recognition.
\newblock In {\em ACM MM}, 2020.

\bibitem{pami1998-sign}
T. Starner, J. Weaver, and A. Pentland.
\newblock Real-time american sign language recognition using desk and wearable
  computer based video.
\newblock {\em TPAMI}, 1998.

\bibitem{seq2seq-Sutskever}
Ilya Sutskever, Oriol Vinyals, and Quoc~V Le.
\newblock Sequence to sequence learning with neural networks.
\newblock In {\em NIPS}, 2014.

\bibitem{taylor2009text}
Paul Taylor.
\newblock {\em Text-to-speech synthesis}.
\newblock Cambridge university press, 2009.

\bibitem{vaswani2017attention}
Ashish Vaswani, Noam Shazeer, Niki Parmar, Jakob Uszkoreit, Llion Jones,
  Aidan~N Gomez, {\L}ukasz Kaiser, and Illia Polosukhin.
\newblock Attention is all you need.
\newblock In {\em NeurIPS}, 2017.

\bibitem{trans-vaswani17attention}
Ashish Vaswani, Noam Shazeer, Niki Parmar, Jakob Uszkoreit, Llion Jones,
  Aidan~N Gomez, {\L}ukasz Kaiser, and Illia Polosukhin.
\newblock Attention is all you need.
\newblock In {\em NeurIPS}, 2017.

\bibitem{viterbi1967error}
Andrew Viterbi.
\newblock Error bounds for convolutional codes and an asymptotically optimum
  decoding algorithm.
\newblock {\em IEEE TIT}, 13(2):260--269, 1967.

\bibitem{signum07}
Ulrich Von~Agris and Karl-Friedrich Kraiss.
\newblock Towards a video corpus for signer-independent continuous sign
  language recognition.
\newblock {\em Gesture in Human-Computer Interaction and Simulation,
  International Gesture Workshop}, 2007.

\bibitem{wang16isolated}
Hanjie Wang, Xiujuan Chai, Xiaopeng Hong, Guoying Zhao, and Xilin Chen.
\newblock Isolated sign language recognition with grassmann covariance
  matrices.
\newblock {\em ACM TACCESS}, 8(4):1--21, 2016.

\bibitem{cslr-CTF-MM}
Shuo Wang, Dan Guo, Wen-gang Zhou, Zheng-Jun Zha, and Meng Wang.
\newblock Connectionist temporal fusion for sign language translation.
\newblock In {\em ACM MM}, 2018.

\bibitem{googletranslation}
Yonghui Wu, Mike Schuster, Zhifeng Chen, Quoc~V Le, Mohammad Norouzi, Wolfgang
  Macherey, Maxim Krikun, Yuan Cao, Qin Gao, Klaus Macherey, et~al.
\newblock Google's neural machine translation system: Bridging the gap between
  human and machine translation.
\newblock {\em arXiv preprint arXiv:1609.08144}, 2016.

\bibitem{xu-etal-2020-clue}
Liang Xu, Hai Hu, Xuanwei Zhang, Lu Li, Chenjie Cao, Yudong Li, Yechen Xu, Kai
  Sun, Dian Yu, Cong Yu, Yin Tian, Qianqian Dong, Weitang Liu, Bo Shi, Yiming
  Cui, Junyi Li, Jun Zeng, Rongzhao Wang, Weijian Xie, Yanting Li, Yina
  Patterson, Zuoyu Tian, Yiwen Zhang, He Zhou, Shaoweihua Liu, Zhe Zhao, Qipeng
  Zhao, Cong Yue, Xinrui Zhang, Zhengliang Yang, Kyle Richardson, and Zhenzhong
  Lan.
\newblock {CLUE}: A {C}hinese language understanding evaluation benchmark.
\newblock In {\em COLING}, 2020.

\bibitem{csl-icme16}
Jihai Zhang, Wengang Zhou, Chao Xie, Junfu Pu, and Houqiang Li.
\newblock Chinese sign language recognition with adaptive hmm.
\newblock In {\em ICME}, 2016.

\bibitem{DPD_icme19}
Hao Zhou, Wengang Zhou, and Houqiang Li.
\newblock Dynamic pseudo label decoding for continuous sign language
  recognition.
\newblock In {\em ICME}, 2019.

\bibitem{stmc-tmm}
Hao Zhou, Wengang Zhou, Yun Zhou, and Houqiang Li.
\newblock Spatial-temporal multi-cue network for sign language recognition and
  translation.
\newblock {\em IEEE TMM}, 2021.

\end{thebibliography}
}

\end{document}